\documentclass[conference]{ieeetran}

\IEEEoverridecommandlockouts    

\usepackage{times}

\usepackage[numbers]{natbib}
\usepackage{multicol}
\usepackage[bookmarks=true]{hyperref}
\usepackage{graphics} 
\usepackage{epsfig} 
\usepackage{mathptmx} 
\usepackage{times} 
\usepackage{amsmath} 
\usepackage{amssymb}  
\usepackage{wrapfig}
\usepackage{multirow}
\usepackage{xcolor}
\usepackage{tablefootnote}
\usepackage{booktabs}
\usepackage{subcaption}
\usepackage{hyperref}
\usepackage{tikz}
\title{\LARGE \bf
All the Feels: A dexterous hand with large-area tactile sensing
}

\author{Raunaq Bhirangi$^{\dagger, 1,2}$, Abigail DeFranco$^{*,1}$, Jacob Adkins$^{*, 1}$, Carmel Majidi$^1$, Abhinav Gupta$^1$, \\ Tess Hellebrekers$^2$ and Vikash Kumar$^{\dagger,2}$
\thanks{$^*$equal contribution}
\thanks{$^\dagger$Correspondence: {\small rbhirang@cs.cmu.edu,vikashplus@gmail.com}}
\thanks{$^{1}$Carnegie Mellon University}%
\thanks{$^{2}$FAIR-MetaAI}%
}

\newcommand{\VK}[1]{}

\newcommand{\RB}[1]{}

\newcommand{\rmd}[1]{}

\newcommand{\change}[1]{{#1}}
\newcommand{\name}{\textit{D'Manus}}
\newcommand{\web}{\href{https://sites.google.com/view/dmanus}{https://sites.google.com/view/dmanus}}

\begin{document}

\maketitle
\thispagestyle{empty}
\pagestyle{empty}

\begin{abstract}
High cost and lack of reliability has precluded the widespread adoption of dexterous hands in robotics. Furthermore, the lack of a viable tactile sensor capable of sensing over the entire area of the hand impedes the rich, low-level feedback that would improve learning of dexterous manipulation skills. This paper introduces an inexpensive, modular, and robust platform - the \name{} - aimed at resolving these challenges while satisfying the large-scale data collection demands of deep robot learning paradigms. Studies on human manipulation point to the criticality of low-level tactile feedback in performing everyday dexterous tasks. The \name{} comes with ReSkin sensing on the entire surface of the palm as well as the fingertips. We also demonstrate the generalizability of tactile models trained with the fully integrated system in a tactile-aware task - bin-picking and sorting. Code, documentation, design files, detailed assembly instructions, trained models, task videos, and all supplementary materials required to recreate the setup can be found on \web
\end{abstract}


\section{Introduction}

Humans routinely operate in unstructured, cluttered environments through a series of surprisingly imprecise, improvised motion. Think about finding the keys hiding at the bottom of your bag, pulling a box from the back of the fridge or finding the steel ladle among the wooden spatulas. While you rely on vision to plan motion at a high level, executing low-level actions involves using a wealth of tactile signals to spatially understand and characterize the environment. The tactile information, combined with natural compliance and underlying motion, enables the effortless dexterity of the human hand. In moving towards robots with human-like sensorimotor abilities, there is a clear need for systems that integrate rich tactile sensing capabilities with dexterous motion.



However, the high dimensionality of dexterous systems also makes them difficult to control. Data-driven methods have emerged as promising approaches to high-dimensional control~\cite{pinto2016supersizing, levine2018learning, Bodnar-RSS-20}, but success with dexterous manipulators has been limited~\cite{andrychowicz2020learning, handa2022dextreme}, and often restricted to simulation~\cite{rajeswaran2017learning, chen2022system}. The contact-rich nature of tasks like in-hand manipulation and tool use makes it difficult for policies learned in simulation to generalize to the real world. Collecting data from real-world interactions, on the other hand, is difficult due to the absence of an affordable, reliable hand that can handle the demands of large-scale data collection. Efforts aimed at developing such hardware have been few and far between~\cite{ahn2020robel, chin2020machine}, largely due to the cost of manufacturing and the lack of reliable sensing and actuation technologies. 

In this work, we leverage recent advancements in rapid prototyping, modular actuation and large-area sensing to present a hand that can make real-world dexterous learning accessible to a wider community of researchers and roboticists. Concretely, our contributions are as follows:

\begin{figure}[t!]
    \centering
    \includegraphics[width=.99\linewidth]{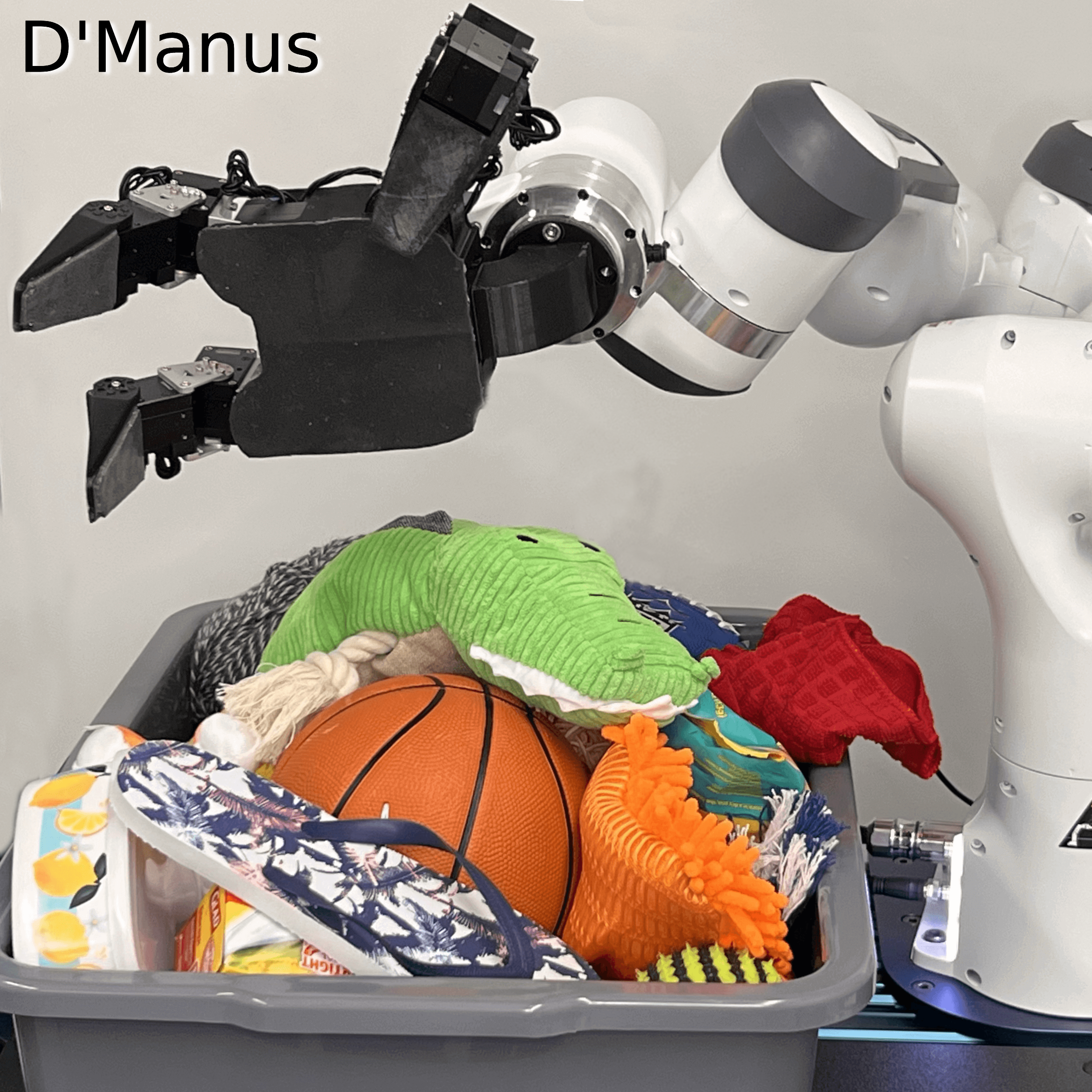}
    \caption{The \name{} -- a low-cost, 10 DoF, reliable prehensile hand with all-over ReSkin~\cite{bhirangi2021reskin} sensing.}
    \label{fig:intro-fig}
    
\end{figure}




\VK{Still missing- our positioning wrt to SOTA. Add that integrated solutions are quite brittle. Add that we have sharp finger tips, which others can't}

\begin{itemize}
    \item We present the \name{} -- an inexpensive, robust prehensile hand geared towards real-world robot learning, complete with a detailed Mujoco-based simulation model for ease of development and prototyping. We rigorously test the hand to withstand long($>$400) hours of operation with no breakages;
    \item We equip the \name{} with customized, integrated ReSkin~\cite{hellebrekers2019soft, bhirangi2021reskin} sensors that provide large-area tactile sensing over the entire surface of the palm and the fingertips, while maintaining sharp fingertips/nails critical for dexterous manipulation;
    \item We demonstrate the caliber of the \name{} system along sensory effectiveness, dexterity, and robustness axes by learning tactile perceptive models for softness and texture identification; and validate their generalizability to unseen objects in a tactile-aware bin sorting task.
    
    
\end{itemize}

\section{Related Work}
\subsection{Dexterous Hands and data-driven learning}
The versatility of the human hand has long inspired a number of efforts aimed at creating similarly capable robotic hands dating back to the early days of robotics ~\cite{mason1985robot, bekey1990control, jacobsen1986design, iberall1997human}. Concurrent work in prosthetics and assistive robotics~\cite{kyberd1994southampton, tomovic1962adaptive, pfeiffer1999shape} has often overlapped with and contributed to research in creating general-purpose robotic hands. More recently, advances in material science and rapid prototyping as well as control algorithms have further pushed the envelope of capable dexterous hands~\cite{piazza2019century}. Since these efforts have primarily been directed towards demonstrating added functionality on human control, they tend to fall short on the scalability, reliability, affordability, and other capabilities required for the prolonged operation demands of robot learning. Despite the recent advancements in data-driven robotics~\cite{pinto2016supersizing, levine2018learning}, robust dexterous platforms capable of meeting the data needs of real-world learning have been few and far in between~\cite{ahn2020robel, zhu2019dexterous, wuthrich2020trifinger}. This has restricted recent investigations with dexterous hands to simulation~\cite{rajeswaran2017learning, chen2022system} or the few researchers who can afford the hardware expense~\cite{andrychowicz2020learning}. \change{The \name{} is an effort aimed at filling this gap with an inexpensive, rigorously tested prehensile hand capable of prolonged operation in contact rich environments.}


\change{Additionally, most recent works aimed at solving dexterous manipulation~\cite{kumar2016learning, nagabandi2020deep, andrychowicz2020learning, zhu2019dexterous} conspicuously use a single exteroceptive sensory modality -- vision~\cite{kopicki2016one, andrychowicz2020learning}. }
\rmd{Additionally, dexterous manipulation evolves in a compact space which is inhibited by static (joint limits) and dynamic constraints (intermittent contact with objects, self collisions, slip) from the hand as well as the object's DoFs. Classical methods ~\cite{cherif1999planning, bicchi2000hands, rus1992dexterous} that explicitly reason over the entire space struggle to scale due to the curse of dimensionality. The rise of deep learning and its promised potential for handling large action spaces has prompted a number of recent works aimed at solving dexterous manipulation using deep neural networks~\cite{kumar2016learning, nagabandi2020deep, andrychowicz2020learning, zhu2019dexterous}. Conspicuously, most of these methods use a single exteroceptive sensory modality -- vision~\cite{kopicki2016one, andrychowicz2020learning}.} Vision provides rich sensory information about the scene and the visual properties of objects, and has been successfully integrated with robot learning frameworks~\cite{pinto2016supersizing, levine2018learning, zeng2020transporter, Bodnar-RSS-20}. However, dexterous tasks are generally contact-rich and require reasoning about contact information that cannot be captured entirely using vision. We posit that the lack of rich tactile information limits a manipulator's ability to effectively perform real-world dexterous manipulation tasks involving force control, flexible objects, and deformable media, particularly with smaller objects which receive degraded visual signals due to occlusions. The \name{} comes with integrated large area sensing that offers a rich tactile sensory modality and extensive spatial coverage suitable for learning such contact-rich manipulation skills.


\begin{figure*}[h!]
    \centering
    \includegraphics[width=.8\textwidth]{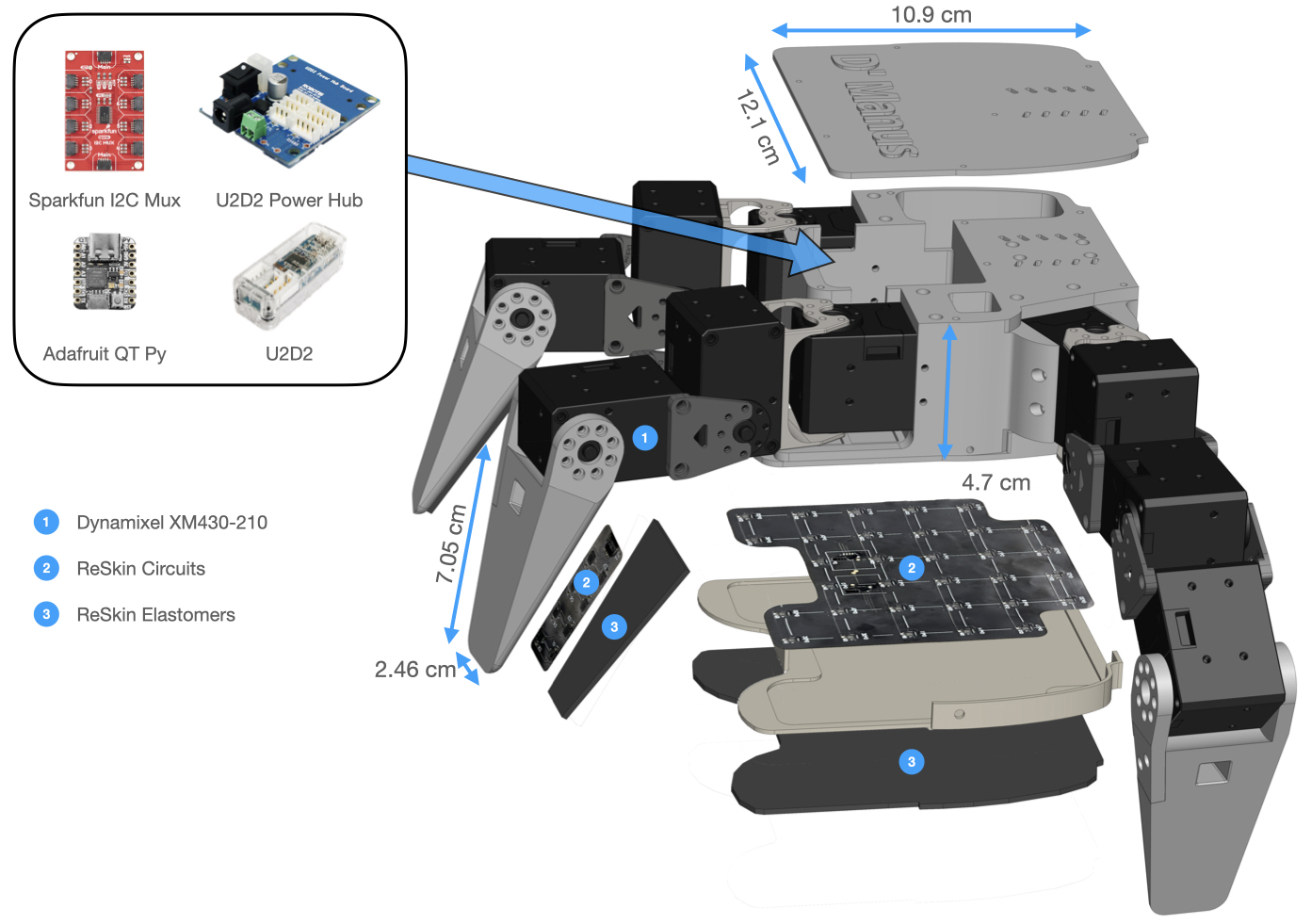}
    \caption{\textbf{Anatomy of the \name{} hand: } The \name{} is actuated at joint level using Dynamixel XM430-210 smart actuators. ReSkin sensors are integrated with the fingertips and the palm. Each fingertip sensor is comprised of 8 magnetometers while the palm sensor consists of 32 magnetometers for a total of 56 magnetometers. Sensor and motor interfacing components are housed in the core of the hand.}
    \label{fig:system-fig}
\end{figure*}
\subsection{Tactile sensing}

The modality of touch has a long history in robotic grasping and manipulation ~\cite{shih2020electronic}. A number of different modalities like capacitive~\cite{cannata2008embedded, hoshi2006large}, resistive~\cite{wettels2008biomimetic}, piezoelectric~\cite{dao2009analysis}, magnetic~\cite{tomo2017covering, hellebrekers2019soft}, audio~\cite{gandhi2020swoosh}, contact microphones~\cite{clarke2018learning,du2022play} and MEMS~\cite{hosoda2006anthropomorphic, van2015learning, odhner2014compliant}-based sensors have been explored as tactile sensing alternatives for robotics. With the recent success of deep learning, especially in computer vision, optical tactile sensors~\cite{yuan2017gelsight, lambeta2020digit, donlon2018gelslim} have emerged as the popular choice of tactile sensor, due to their high resolution as well as their compatibility with popular neural architectures (CNNs) for processing signals. Most of these solutions, however, have limitations that significantly impede their ability to serve as effective sensors for capable hands, which have strict requirements in terms of sensing, space, cost and robustness. Some \VK{Value.  Try to be precise. Group them into some categories and talk about the category. Or talk more generally as them being a set of challenges, not all can be satisfied at once} of these sensors are bulky~\cite{yuan2017gelsight, lambeta2020digit} or need direct electrical connections between the circuitry and the interface~\cite{hosoda2006anthropomorphic, dao2009analysis, wettels2008biomimetic}, resulting in design constraints that compromise on the manipulation abilities of the hand. Some others \VK{same as above} are either expensive~\cite{wettels2008biomimetic} or difficult to fabricate~\cite{cannata2008embedded, hoshi2006large, donlon2018gelslim} and cannot be easily replaced, making them less suitable for large-scale data collection given the inevitable wear-and-tear that comes from frequent contact with a wide variety of objects. Yet other alternatives that are affordable and have suitable form factors for dexterous hands tend to lack the resolution, shear sensing~\cite{odhner2014compliant, heyneman2016slip, mittendorfer2015realizing} \VK{and surface coverage~\cite{donlon2018gelslim}} required for fine-grained control. Manufacturing, cost, and reliability challenges only escalate with larger area sensing systems, such as the MIT Glove ~\cite{sundaram2019learning}, hex-o-skin~\cite{mittendorfer2015realizing, dean2019whole}, and uSkin~\cite{funabashi2019morphology,funabashi2022multi} among others~\cite{cannata2008embedded, hoshi2006large, odhner2014compliant, heyneman2016slip}.

\change{The class of works that come closest to our proposition are~\cite{funabashi2019morphology, funabashi2022multi} that use $uSkin$ sensors to sensorize a dexterous hand and demonstrate application in object classification and manipulation tasks. However, $uSkin$ uses macro-scale magnets embedded in elastomer as the sensing interface, which involves complex design to avoid crosstalk between magnetometers~\cite{tomo2017covering, tomo2018new}, is bulky, and creates an external magnetic field which can interfere with the environment. To counteract all of these problems, we turn to ReSkin~\cite{hellebrekers2019soft, bhirangi2021reskin}, which is similar to uSkin in sensing principle, but differs critically in the use of magnetic microparticles instead of macro-sized magnets. ReSkin offers the \name{} a number of key advantages as a dexterous hand for robot learning, namely, (a) favorable form factor: ReSkin can be much thinner($\sim$2mm) than its closest alternatives ($>$5mm) enabling sharp fingernails critical to dexterous manipulation, (b) cost and replaceability: ReSkin is easily replaceable~\cite{bhirangi2021reskin} and costs 50x lesser per sensor ($\sim \$20$) than alternatives like uSkin ($\sim \$1000$), and (c) wear resistance: the absence of a hard-soft interface within the elastomer significantly improves the durability of ReSkin ~\cite{bhirangi2021reskin} and, as a result, the \name{}.}

\section{Platform and System Details}
The \name{} is a low-cost, reliable prehensile robotic hand with immersive tactile sensing over its larger contact surfaces, i.e. the palm and fingertips as anatomized in Fig. \ref{fig:system-fig}. To benefit the community and facilitate adoption, \name{} is released as an open-sourced manipulation platform\footnote{CAD models, bill of materials, circuit designs, assembly and setup instructions can be found on \web}. In this section, we detail the features and properties of the system.

\begin{table}[h!]
    \centering
    \change{\begin{tabular}{p{0.6\linewidth} p{0.2\linewidth}}
         Component & Cost \\\toprule
         ReSkin Circuits & \\
         \hspace{3mm} Boards & \$ 38.50\\
         \hspace{3mm} Assembly & \$ 149.60\\
         \hspace{3mm} Parts & \$ 56.00\\
         Magnetic Microparticles & \$ 5.50\\
         Smooth-On DragonSkin-10 NV & \$ 5.50\\
         3D printed components & \$ 50.00\\
         Machined components & \$ 200.00\\
         Sparkfun I2C Mux & \$ 12.95 \\
         Adafruit QT Py & \$ 7.50 \\
         U2D2 & \$ 32.10 \\
         U2D2 Power Hub & \$ 19.00 \\
         Dynamixel XM430-210 motors & \$ 2899.00 \\\midrule
         Total & \$ 3475.65 \\\bottomrule
    \end{tabular}}
    \caption{Cost breakdown for the D'Manus}
    \label{tab:cost-breakdown}
\end{table}

\subsection{The Hand: Construction and Interfacing}
\label{hand_details}
The \name{} hand is a three-fingered, 10-DoF hand -- each finger has three degrees of freedom, with a fourth DoF for the thumb. The hand is actuated at joint-level via Dynamixel smart actuators. A 12V power supply is used to power the hand and a USB-serial bus is used for communication between the hand and a control computer. Non-actuated elements of the hand like the palm and the fingertips are 3D printed and the actuators are daisy chained. This allows the \name{} to be easily customized and assembled while maintaining a low price point(\$3500), as detailed in Table \ref{tab:cost-breakdown}. The hand can be made compatible to be mounted on any robot arm or wrist attachment of choice using a simple 3D printed adaptor. While we experimented with versions of the platform with up to 16 DoFs, we converged on the 10 DoF \name{} as it strikes a balance between dexterity, cost, robustness, weight, and size. 

\subsection{Large-area Exteroceptive Sensing: ReSkin}
\label{sec:reskin}
We use ReSkin~\cite{hellebrekers2019soft, bhirangi2021reskin} to endow the hand with large-area exteroceptive tactile sensing. ReSkin uses a magnetic elastomer interface with magnetometer circuits underneath to detect deformation. Drawing from~\cite{bhirangi2021reskin}, we scale the sensor circuits and the skins to the size of the palm and the fingertips while maintaining a thickness of 2mm for the skins. Each fingertip sensor is comprised of 8 magnetometers, while the palm sensor consists of 32 magnetometers for a total of 56 magnetometers. The signal from each magnetometer is the 3-axis magnetic flux density. Where our approach deviate significantly from ~\cite{bhirangi2021reskin} is an improved fabrication procedure for the magnetic elastomer skins used in this work. We create similar molds scaled to the size of the fingertips and palm, but instead of curing the elastomers under a magnetic field using grids of magnets, we use a pulse magnetizer. The skins are first allowed to cure at room temperature without interfering magnetic fields, and then magnetized using a pulse magnetizer with a 4 Tesla (40 kOe) impulse. This change results in two improvements: (a) stronger signal strength (2-3x) for the same deformation, and (b) ease and scalability of fabrication by eliminating the use of magnetic grids that scale poorly with the size of the skin. Data from the sensors is streamed to the control computer via USB through a microcontroller + I2C mux. Fig. \ref{fig:system-fig} illustrates the construction of the hand and how it integrates with the tactile sensors. A highlight of this design is also the large sensorized area of the palm ($\sim$ 11 cm x 12 cm) which facilitates stable power grasps and provides a base with force feedback for objects during in-hand manipulation tasks.

\subsection{Control and Proprioceptive Sensing}
\label{subsec:sensing-control-reliability}
In addition to exteroceptive feedback from the skin, the criticality of proprioceptive feedback from finger muscles in dexterous manipulation for humans is well-documented~\cite{johansson1996sensory}. Recent work in robotic manipulation~\cite{righetti2014autonomous, xia2022review} has effectively leveraged intrinsic sensor feedback for dexterous control. To enable closed loop manipulation strategies with strong sensory feedback, the \name{} also comes with a range of proprioceptive sensing capabilities at the actuated joints, as listed in Table \ref{tab:dmanus-options}.

Control strategies for manipulators lie on a spectrum between position/velocity control and force control~\cite{hogan1985impedance, mason1981compliance}. When interaction forces are negligible, position control enables more precise control of the end effector, while velocity control allows for smoother movements. On the other hand, constraints in the environment and frequent interaction forces lend themselves better to force control or ``compliant" strategies~\cite{mason1981compliance}. Over the years, there have been unifying hybrid approaches such as impedance and admittance control~\cite{hogan1985impedance}. The use of Dynamixel smart actuators afford the \name{} a number of control modes as outlined in Table \ref{tab:dmanus-options}, allowing operational flexibility for end user applications\footnote{PWM mode allows for force control. Current mode allows for hybrid position and force control}.

 



\begin{table}[ht]
    \centering
    \begin{tabular}{p{0.3\linewidth}|p{0.58\linewidth}}
         Property & Options \\\hline
         Control & Position, Velocity, Current, PWM \\
         Proprioceptive Sensing & Position, Velocity, Current, Realtime tick, Trajectory, Input Voltage \\
         Exteroceptive Sensing & ReSkin (30 Hz) \\
         Limits & Position, Velocity, PWM, Current \\
         Baudrate & 9600 bps $\sim$ 4.5 Mbps \\\hline
    \end{tabular}
    \caption{Operational Details for the D'Manus}
    \label{tab:dmanus-options}
\end{table}

\subsection{Software}
\begin{figure}[h!]
    \begin{center}
    \includegraphics[width=0.5\linewidth]{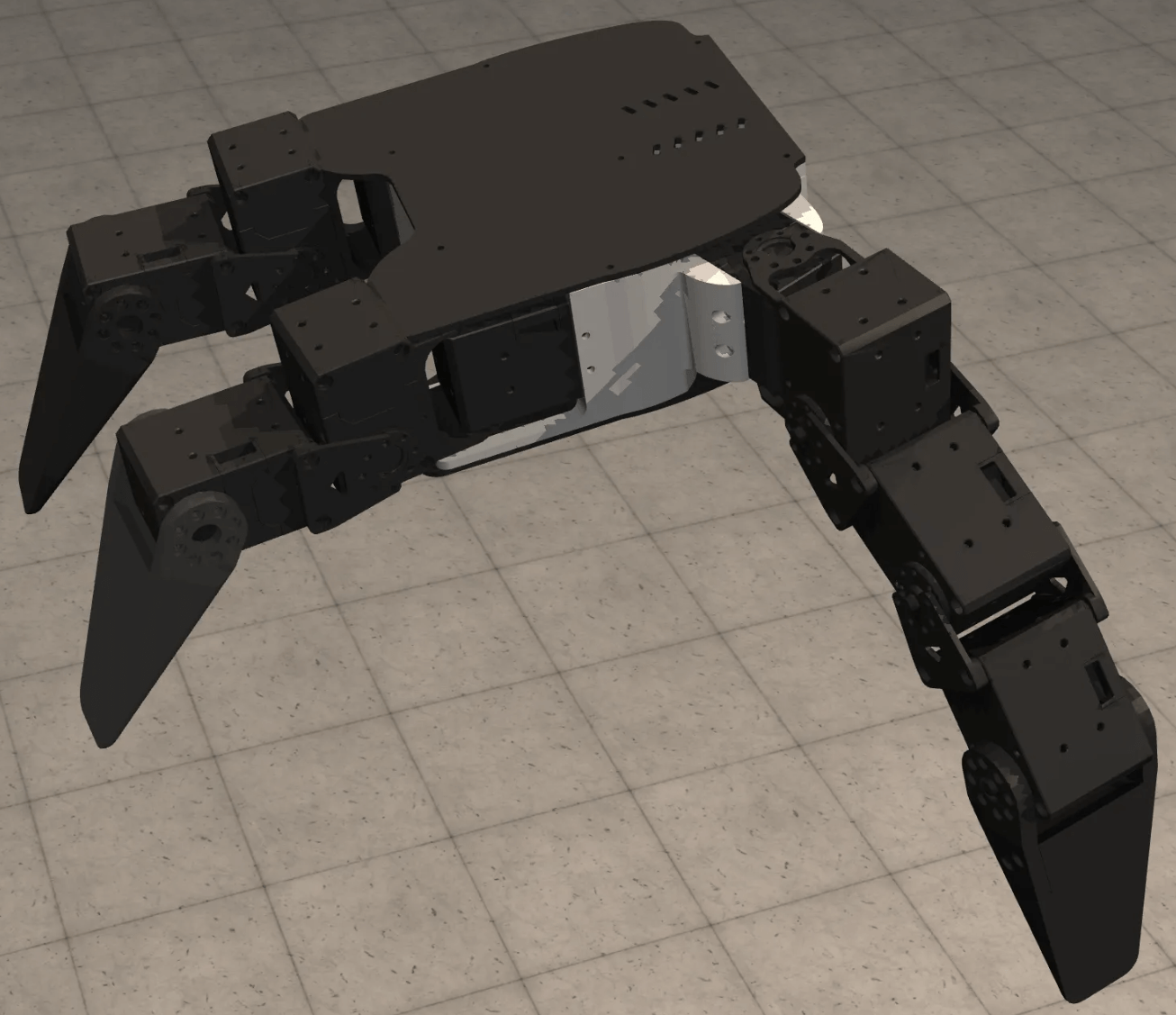}
  \end{center}
  \caption{Simulated \name{}}
  \label{fig:dmanus_sim}
\end{figure}

The \name{}'s software package includes a python driver that exposes all the operational modalities outlined in Table \ref{tab:dmanus-options}, a detailed simulation model of the \name{} based on MuJoCo (Figure \ref{fig:dmanus_sim}), and an approximate data-driven model of ReSkin sensors intended for experimental prototyping. The software has been structured for ease of prototyping in simulation and seamless transfer from simulation to the real hardware.

\section{Experiments}

The \name{} is designed to sustain and support long hours of contact-rich interactions and data collection with minimal, easily fixable breakages. Such robustness allows the \name{} to be used for long durations in a real-world robot learning setup similar to the systems demonstrated in \textit{ROBEL} ~\cite{ahn2020robel}. We structure our evaluations investigating effectiveness of \name{} as a testbed for real world robot learning along various axes --
\begin{enumerate} 
    \item \textit{Dexterity}: In section \ref{sec:dexterity}, we evaluate \name's  prehensile ability by subjecting it to a variety of objects and grasping scenarios.
    \item \textit{Tactile Perception}: In sections \ref{sec:material-identification}, \ref{sec:softness-texture-id}, we validate the discriminative ability of extensive tactile sensing by subjecting the \name{} to perform material, softness, and texture identification purely based on surface properties.
    

    \item \textit{Perceptive Generalization}: We demonstrate generalization of learned tactile models for softness and texture identification to unseen objects in section \ref{sec:softness-texture-id}, and unseen tasks in section \ref{sec:bin-picking}, to substantiate the stationarity and richness of ReSkin data.
    
    \item \textit{Integrated system}: We also corroborate the capabilities of the integrated \name{} in section \ref{sec:bin-picking} by exposing it to unseen, real-time interactions in a bin picking setup and demonstrating automated bin sorting purely from tactile information (no visual inputs). 
    
    \item \textit{Robustness}: Finally, in section \ref{sec:robustness}, we outline \name's endurance and resilience towards extended periods of interaction rich operation.
\end{enumerate}

While the \name{} can be mounted on any robot arm using a 3D printed attachment, we used Franka Emika Panda robot (integrated using the \texttt{polymetis} drivers~\cite{Polymetis2021}) for all our experiments. The experiments presented in this paper are performed in a tabletop setting. Neural network models presented in the following sections are are trained on a single GPU (NVIDIA GeForce GTX 1080 Ti). 

In the following section, we elaborate on the data collection and modeling choices for our learned tactile perception models before presenting experimental results in more detail.

\section{Tactile Perception: Data Collection and Modeling}
To exemplify the tactile sensing capabilities afforded to the \name{} by integrated large-area ReSkin sensing, we learn generalizable tactile perception models by training on ReSkin interaction data from a variety of objects. In this section, we detail the data collection setup and the modeling frameworks used to build these perception models.

\subsection{Data Collection}
\begin{figure}[ht]
    \centering
    \includegraphics[width=.9\linewidth]{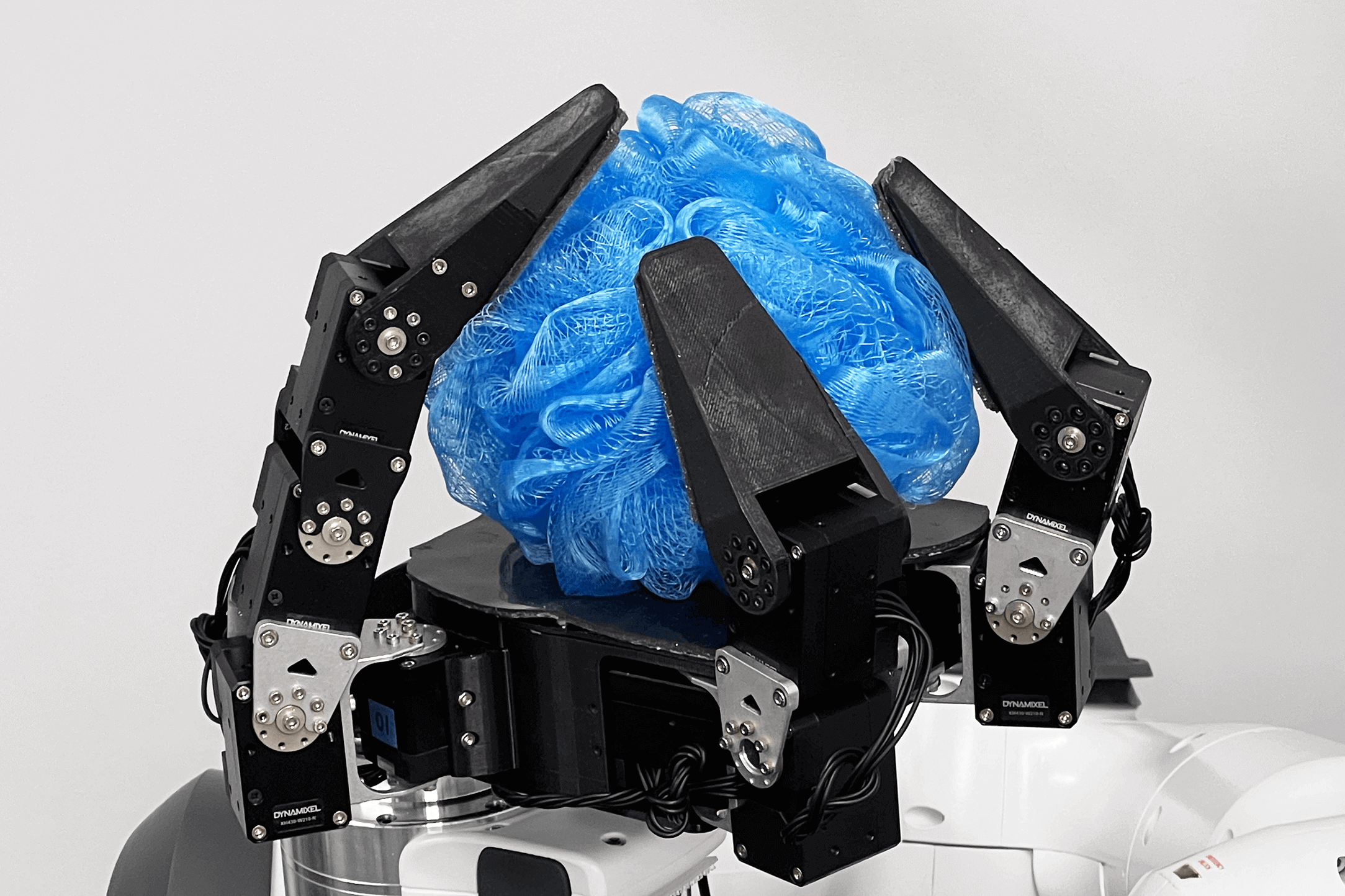}
    \caption{\textbf{Data collection setup:} Tactile data is collected by placing the object on the palm and executing a human-scripted interaction policy for motor babble.} 
    \label{fig:data-collection-setup}
\end{figure}

To collect tactile interaction data, we fix the \name{} such that the palm is facing upwards as shown in Fig. \ref{fig:data-collection-setup}. For every object, we collect several trajectories of interaction data by placing it on the palm and executing a noisy, scripted motor babbling policy (at 30 Hz control frequency) for 10 seconds. While it is possible to use more sophisticated data collection policies such as a learned tactile exploration policy ~\cite{martinez2017active}, we found that simple motor babbling provided sufficient data diversity for training our models, and demonstrating generalization to unseen objects and tasks.

\begin{figure}[ht]
    \centering
    \includegraphics[width=.99\linewidth]{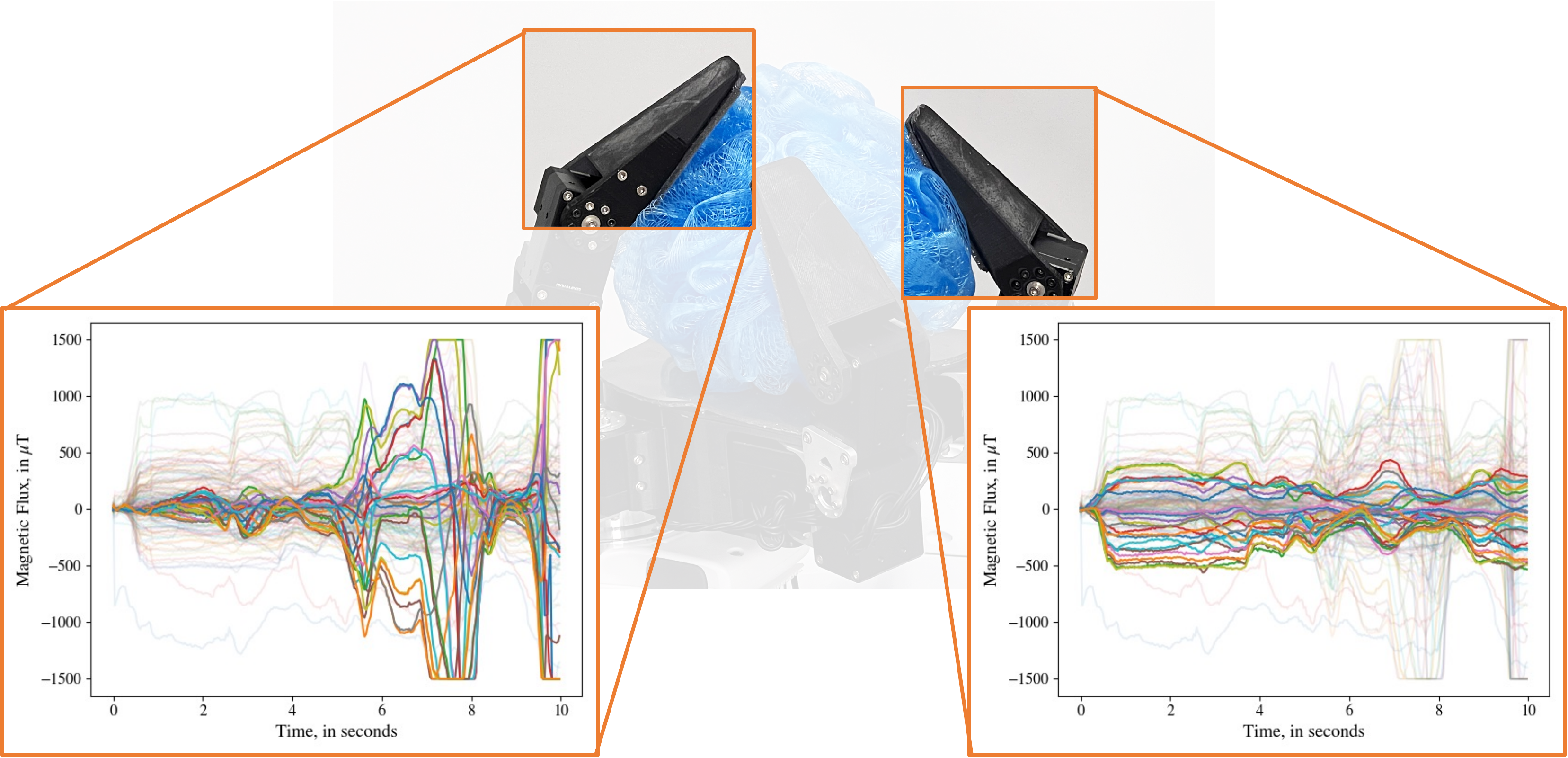}
    \caption{\textbf{Sample ReSkin data:} Visualization of tactile data from two of the fingers while interacting with the loofah in Fig. \ref{fig:data-collection-setup}.} 
    \label{fig:data-viz}
\end{figure}

\label{subsec:data-collection}
As the interaction policy is executed, ReSkin data from the fingertips and the palm is streamed to the control computer at every time step. Each frame of data consists of 3-axis magnetic flux measurements for each of the 56 magnetometers enumerated in Fig. \ref{fig:system-fig}. Sample data from an interaction trajectory can be seen in Fig. \ref{fig:data-viz}. A more dynamic visualization of the raw data can be found in the accompanying video.


\subsection{Model Learning}
\label{sec:model-learning}

\begin{figure}[h!]
    \centering
    \includegraphics[width=0.9\linewidth]{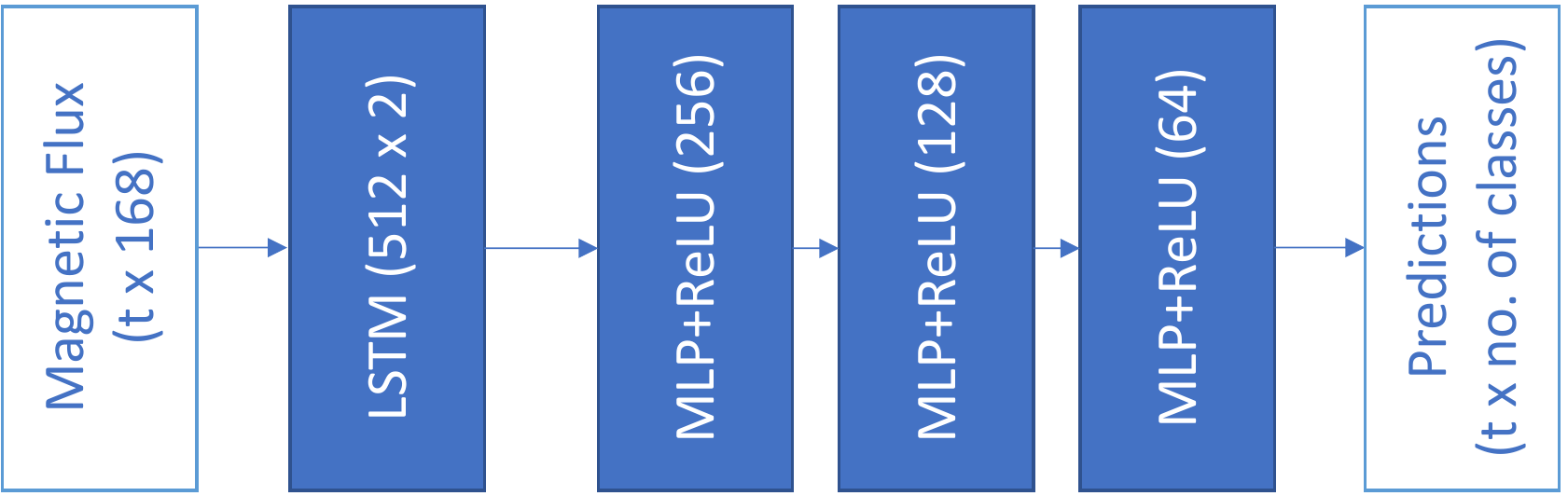}
    \caption{Model architecture}
    \label{fig:model_arch}
\end{figure}

Vision-based tactile sensors~\cite{yuan2017gelsight, lambeta2020digit, donlon2018gelslim} have naturally leveraged convolutional neural networks (CNNs) as a backbone for processing tactile information, as the signal is fundamentally visual RGB values. In contrast, the electromagnetic signals of ReSkin have much less redundancy and have relatively lower dimensionality. In order to allow the learning algorithm to pick from a larger class of functions, we choose to use fully connected multilayer perceptrons(MLPs) as building blocks for the neural architecture used to process ReSkin signal. Furthermore, contact information from interaction data is naturally sequential, and our model architecture must be capable of leveraging temporal correlations in the data. To ensure this, we use a recurrent neural architecture, an LSTM, at the base of our model. The neural architecture used in this work, as shown in Fig. \ref{fig:model_arch}, consists of an LSTM with 2 hidden layers with 512 nodes each, followed by 3 fully connected layers of sizes 256, 128 and 64 with ReLU activations, and an output layer that represents a categorical distribution over classes. All the models in the experiments presented below are classification models trained using the standard cross-entropy loss, and map a sequence of magnetic flux vectors to probability of classes. Predictions are made at every timestep of the interaction trajectory, and the target label corresponds to the task-specific label of the object being interacted with. Each frame of tactile data from ReSkin consists of concatenated signals from each of the magnetometers, resulting in a 168-dimensional magnetic flux vector. 

\begin{figure*}[tbh]
    \centering
    \includegraphics[width=.9\textwidth]{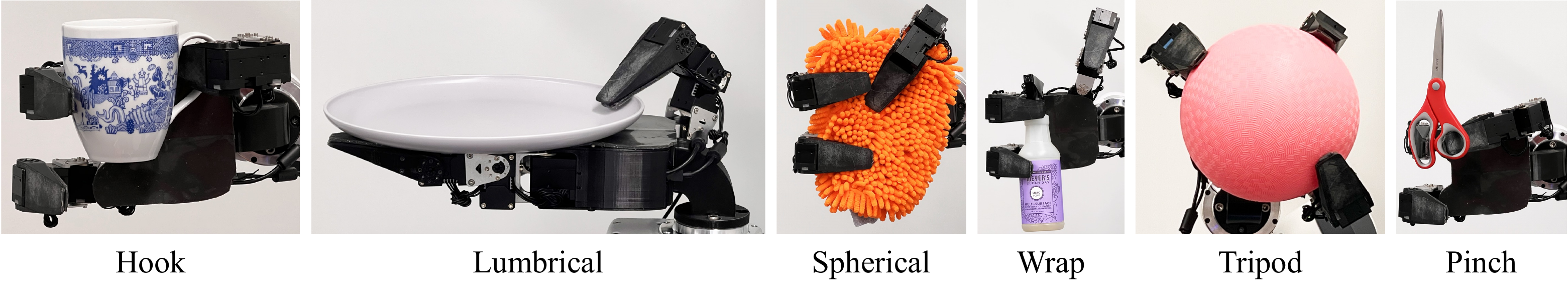}
    \caption{Illustration of the \name{} grasping different objects using a variety of grasps~\cite{yang2015grasp, feix2015grasp} }
    \label{fig:dmanus-grasps}
\end{figure*}

We also provide an ablation in Appendix \ref{ablation_details} where we replace the LSTM with an MLP that takes stacked frames of ReSkin data as input. Details of parameter sweeps over model parameters can be found in Appendix \ref{ablation_details}.

\section{Results}
\label{sec:results}

\subsection{Dexterity of the \name{}}
\label{sec:dexterity}
We qualitatively demonstrate the dexterous capabilities of the \name{} (Fig. \ref{fig:dmanus-grasps}) using interactions with everyday objects. We observe that the \name{} is effective at grasping and (in-hand as well as hand-arm) manipulation of day-to-day objects. Its abilities, however, are somewhat restricted for in-hand manipulation of small objects (e.g. counting coins on palm). This is in accordance with the dexterity and robustness trade-off we made and detailed in section \ref{hand_details}.

\subsection{Tactile Perception: Material Identification}
\label{sec:material-identification}
\begin{figure}[h!]
    \centering
    \includegraphics[width=\linewidth]{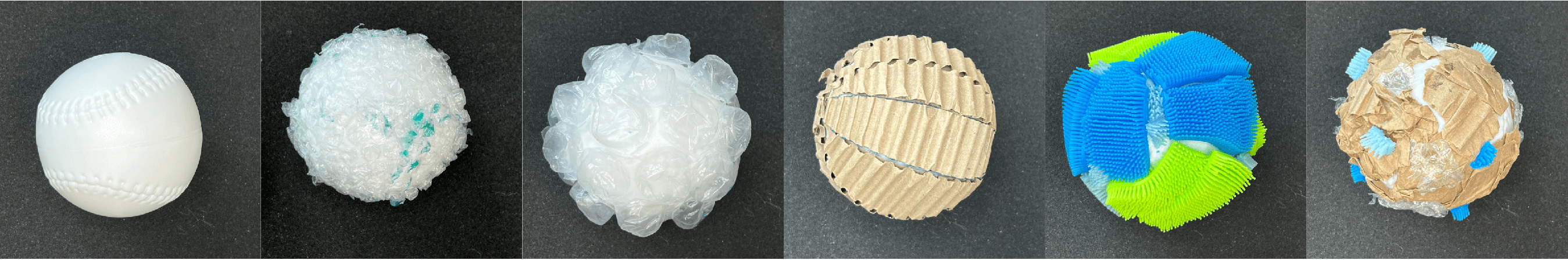}
    \caption{\textbf{Material coverings for Material Identification Task:} Uncovered, small bubble wrap, large bubble wrap, corrugated cardboard, silicone sponge and combination of materials}
    \label{fig:expt1-balls}

\end{figure}

To establish the effectiveness of tactile perceptual capabilities, we task \name{} to leverage its tactile signals to classify objects merely based on their surface properties in a simple material identification task. The idea is to demonstrate that we can build tactile models capable of quantifying the differences between the tactile signature of different materials as obtained by the \name{}. We pick a set of six identical balls, each with a different outer covering -- small bubble wrap, large bubble wrap, corrugated cardboard, silicone sponge, a combination of all these materials, and no covering material -- as shown in Fig. \ref{fig:expt1-balls}, and collect interaction data as described in section \ref{subsec:data-collection}. We ensure that our models can only rely on surface properties by using balls of identical shape and size. We collect 35 trajectories for each ball and use a 30-5 train-validation split. 

\VK{Here is an interesting question -- Since this model is shape agnostic. Will it be able to classify these materials on any other shape? Say cubes?}

We train classification models to learn to predict a probability distribution over the six materials from tactile interaction data, as detailed in section \ref{sec:model-learning}. Our models show a 71.24\% validation accuracy on the 30 held-out trajectories (5 per material) described above, as shown in Table \ref{tab:results}. The success of our models to distinguish between the six balls confirms the discriminability of the tactile interaction data obtained by the \name{}.




\begin{table}[h!]
    \centering
    \begin{tabular}{c c}
         \toprule
         Task & Validation accuracy \\\hline
         Material Identification \rule{0pt}{2.5ex} & 71.24\% \\ \midrule
         Softness Classification & 76.17\% \\
         Texture Classification & 59.03\% \\\bottomrule
    \end{tabular}
    \caption{The \name{} can distinguish between different materials purely using tactile feedback (Sec. \ref{sec:material-identification}). Further, models trained for softness and texture classification generalize to interactions with unseen objects (Sec. \ref{sec:softness-texture-id}). \VK{I'd recommend listing all 3 texture models here, without them the sec 6.c will be very hard to understand.} \VK{Can we also add a few lines in the section how we are validating our models and what validation accuracy stands for.}}
    \label{tab:results}
\end{table}

\subsection{Perceptive Generalization: Softness and Texture Identification}
\label{sec:softness-texture-id}

Having verified the distinguishability of sensory signals of \name{}, we shift our focus to the consistency of the sensory signal across different objects and scenarios. To investigate this, we develop tactile perception models for softness and texture identification and demonstrate its generalization to unseen objects. For generalization, it is imperative that the (a) \name's sensors capture overall surface characteristics from interaction, (b) sensory signatures do not drift, and stay consistent over time, and (c) tactile perception models are effective outside the training environment. To demonstrate that our tactile perception is robust to such drift, we collect training data over an extended period of time (a few days). 

\begin{figure*}[bth]
    \centering
    \includegraphics[width=.95\textwidth]{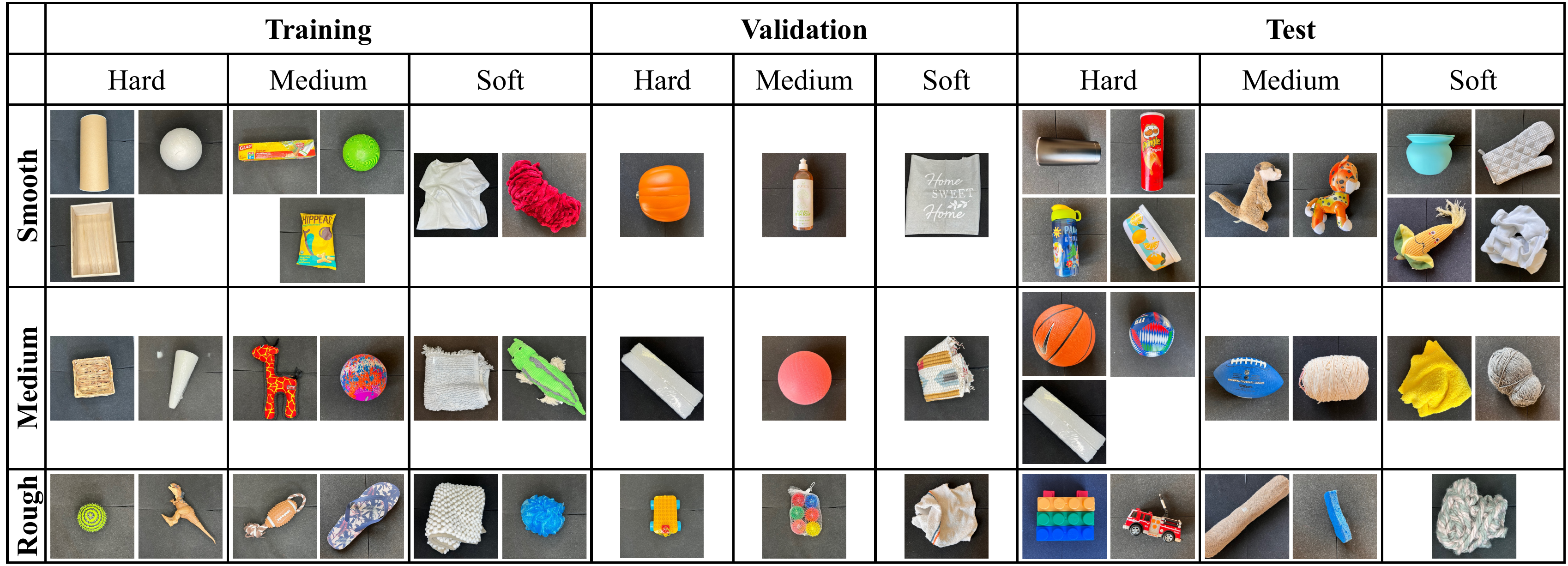}
    \caption{Datasets used for Softness and Texture Identification Models}
    \label{fig:dataset}
\end{figure*}

To learn tactile identification models, we would need quantifiable descriptions of surface characteristics. We create a three-point scale to quantify softness -- \textit{Hard}, \textit{Medium}, \textit{Soft} -- as well as texture -- \textit{Smooth}, \textit{Medium}, \textit{Rough}. We manually assign softness and texture labels to over 50 objects by consensus among the authors \textbf{before} starting the study. We use a set of 20 training objects and 9 validation objects for these tasks. The full set of objects and their split can be found in Fig. \ref{fig:dataset}. The corresponding datasets are created by collecting 15 trajectories of tactile interaction data for each of the training objects and 5 trajectories for each of the validation objects. We now use the collected training data to train softness and texture identification models and examine their generalizability to unseen objects.


\subsubsection{Softness Classification}
\label{sec:softness-classification}
We train a classification model as described in section \ref{sec:model-learning} to predict softness categories, and the results are presented in Table \ref{tab:results}. We see that our models are able to successfully classify objects on the softness scale defined above. 




\subsubsection{Texture Classification}
\label{sec:texture-classification}
Training a model for texture classification analogous to softness classification has low generalization on validation set. This is understandable as object's texture is dependent on its softness as well. Texture, in this context, refers to spatial discontinuities in the force profile felt by the interface in contact with the object. Gradation of texture therefore is difficult to maintain across different softness categories. For example, think about a soft, neatly folded T-shirt -- while this folded T-shirt can be said to have smooth texture, the same T-shirt clumped into a ball will have a number of folds that could make it feel highly textured. On the other hand, for a harder object like the spiky ball in the bottom left corner of Fig. \ref{fig:dataset}, the texture remains consistent due to its structural integrity. Furthermore, the interaction force profile for the ball will tend to have more spatial discontinuities than softer textured objects. Thus, object texture is difficult to quantify and analyze independently of softness properties. 

To get around this problem, we train softness-conditioned texture identification models -- individual \textit{Softness-Conditioned} models trained on the subset of the training data corresponding to the softness category. The accuracy presented in Table \ref{tab:results} is the mean accuracy over the three softness categories. Appendix \ref{ablation_details} contains an ablation with the different types of models we trained, and detailed comparisons between them.

\VK{I'll recommend moving all there RNN texture models to the main paper. Its hard to follow the description here without them. Also you are refer to the models using the names. You tend to use third person language that makes its a bit hard to follow}.

A highlight of \name{} design, as outlined in section \ref{sec:reskin}, is the large sensorized area of the palm. To corroborate this design choice, we train standalone softness and texture identification models corresponding to each individual fingertip and the palm. We evaluate the performance of these standalone models and present a comparison in Table \ref{tab:component-models}. The performance on the individual finger models is significantly lower than the palm as well as the full prediction model. While some of this discrepancy can be attributed to the fingers losing contact with the object during parts of the interaction trajectory, strong performance of other models strongly emphasizes the benefits of large area sensing available on \name{}.

\begin{table}[h!]
    \centering
    \begin{tabular}{c c c}
         \toprule
         Component  & Softness Accuracy & Texture Accuracy\\
         \hline
         \rule{0pt}{3ex}Finger 1  & 56.27\% & 49.77\%\\
         Finger 2  & 48.48\% & 46.99\%\\
         Finger 3  & 59.59\% & 50.50\%\\
         Palm & 74.31\% & 54.20\%\\
         All & 76.17\% & 59.03\% \\\bottomrule
    \end{tabular}
    \caption{Comparison of models trained using data from different components of the hand. }
    \label{tab:component-models}
    
\end{table}

Having successfully inspected and clarified the generalizability of our perceptive models across objects and time, we can assert the strength of tactile perceptual capabilities if \name{}. To further substantiate this argument, and demonstrate the working of the integrated system in an unseen environment, we deploy the learned softness and texture identification models in a tactile-aware bin sorting task in the following section.




\subsection{Tactile Bin Sorting}
\label{sec:bin-picking}
As our final experiment, we access the ability of our trained models to generalize to realistic environments and tasks. For this evaluation, we pick a cluttered bin sorting experiment. We attempt to pick objects from a cluttered bin and sort them according to softness and texture from tactile signals as a test of the generalizability of our learned models. We start with a cluttered bin containing a variety of objects as shown in Fig. \ref{fig:intro-fig}. The robot samples a random location above the bin and reaches down into the bin. Once the magnitude of the ReSkin signal from the \name{} exceeds a certain threshold, the hand stops moving and executes a grasp. If it successfully grasps an object, we predict the softness and texture of the grasped object, and place it in the corresponding labelled box. We then replace it by adding a new object to the bin and the process is continued. Over 20 successful grasps of different objects, our models achieve a prediction accuracy of 65\% (ie. 13 objects) on both softness and texture prediction, confirming the ability of our models to extend to unseen tasks and environments in the real world. Further, it is worth noting that the tactile models trained with the hand upright are able to generalize to this setting where the hand is primarily operated in a downward-facing configuration. Through this experiment, we validate the ability of the integrated \name{} system as well the generalizability of our models in performing tactile-rich tasks in unseen, real-world environments.

\subsection{Robustness and Reliability}
\label{sec:robustness}
Amongst various versions of the platform, we have logged over 10,000 hours of operational time over the course of 12 months in 3 different locations with a total of 5 breakages. These breakages consisted of three motor failures, one 3D printed part failure, and operational deterioration of wires -- all of which were repaired in-house within 30 minutes by non expert users. The version of the platform being released has significantly benefited from aggressive real world testing of prior versions. The specific copy of the \name{} we are presenting experimental results on has been used for over 400 hours over the last 8 months with no breakages, corroborating our claims about the robustness and reliability of this system to operate for long durations for real-world learning in contact-rich robotic tasks.

\section{Conclusions and Limitations}
We present the \name{} -- a low-cost, 3D printable, prehensile robotic hand geared towards robot learning. The hand comes with multiple actuation modes, proprioceptive sensing abilities as well as ReSkin-based large-area tactile sensing. We demonstrate the dexterity of this platform in grasping a variety of objects. To exemplify the utility of the large-area sensing, we validate the discriminability of the tactile signal by learning models for material identification as well as category-level softness and texture identification. Further, we illustrate the transferability of learned tactile models to unstructured, real-world envirionments through a touch-based bin picking and sorting task. \change{The design, assembly and setup instructions have all been open-sourced to facilitate adoption and integration by the community}




\noindent\textbf{Limitations}: While we validate the tactile capabilities of the hand, we would like to fully evaluate the dexterous capabilities of the hand by integrating tactile sensing into a dexterous policy learning setup in future work. \change{The global semiconductor shortage limited the number of magnetometer chips we were able to integrate into this system. As a result, the system lacks sensing on the phalanges, ie. the surface of the motors, as well as the sides and backs of the fingertips. The same design principles outlined in section \ref{sec:reskin} would make this a simple extension of the present version and will be addressed in subsequent work.} We also believe that unlocking the full potential of all-over tactile sensing requires the integration of other sensory modalities like vision and audio, allowing the robotic system richer sensory inputs to solve complex dexterous tasks. Finally, another limitation of this work is the lack of quantitative comparisons to other existing platforms due to limitation in access and the high cost involved in such a pursuit. We hope that our effort in open-sourcing \name{} and its low cost will help with comparative evaluations with our system.


\section*{ACKNOWLEDGMENTS}

The authors thank Vani Sundaram for her help with fixing ReSkin circuits, and all the members of FAIR Pittsburgh, AGI-Labs, and SML at CMU for their valuable inputs.

\newpage

\bibliographystyle{ieeetr}
\bibliography{citations}

\begin{thebibliography}{10}

\bibitem{pinto2016supersizing}
L.~Pinto and A.~Gupta, ``Supersizing self-supervision: Learning to grasp from
  50k tries and 700 robot hours,'' in {\em 2016 IEEE international conference
  on robotics and automation (ICRA)}, pp.~3406--3413, IEEE, 2016.

\bibitem{levine2018learning}
S.~Levine, P.~Pastor, A.~Krizhevsky, J.~Ibarz, and D.~Quillen, ``Learning
  hand-eye coordination for robotic grasping with deep learning and large-scale
  data collection,'' {\em The International journal of robotics research},
  vol.~37, no.~4-5, pp.~421--436, 2018.

\bibitem{Bodnar-RSS-20}
C.~Bodnar, A.~Li, K.~Hausman, P.~Pastor, and M.~Kalakrishnan, ``Quantile qt-opt
  for risk-aware vision-based robotic grasping,'' in {\em Proceedings of
  Robotics: Science and Systems}, (Corvalis, Oregon, USA), July 2020.

\bibitem{andrychowicz2020learning}
O.~M. Andrychowicz, B.~Baker, M.~Chociej, R.~Jozefowicz, B.~McGrew,
  J.~Pachocki, A.~Petron, M.~Plappert, G.~Powell, A.~Ray, {\em et~al.},
  ``Learning dexterous in-hand manipulation,'' {\em The International Journal
  of Robotics Research}, vol.~39, no.~1, pp.~3--20, 2020.

\bibitem{handa2022dextreme}
A.~Handa, A.~Allshire, V.~Makoviychuk, A.~Petrenko, R.~Singh, J.~Liu,
  D.~Makoviichuk, K.~Van~Wyk, A.~Zhurkevich, B.~Sundaralingam, {\em et~al.},
  ``Dextreme: Transfer of agile in-hand manipulation from simulation to
  reality,'' {\em arXiv preprint arXiv:2210.13702}, 2022.

\bibitem{rajeswaran2017learning}
A.~Rajeswaran, V.~Kumar, A.~Gupta, G.~Vezzani, J.~Schulman, E.~Todorov, and
  S.~Levine, ``Learning complex dexterous manipulation with deep reinforcement
  learning and demonstrations,'' {\em arXiv preprint arXiv:1709.10087}, 2017.

\bibitem{chen2022system}
T.~Chen, J.~Xu, and P.~Agrawal, ``A system for general in-hand object
  re-orientation,'' in {\em Conference on Robot Learning}, pp.~297--307, PMLR,
  2022.

\bibitem{ahn2020robel}
M.~Ahn, H.~Zhu, K.~Hartikainen, H.~Ponte, A.~Gupta, S.~Levine, and V.~Kumar,
  ``Robel: Robotics benchmarks for learning with low-cost robots,'' in {\em
  Conference on robot learning}, pp.~1300--1313, PMLR, 2020.

\bibitem{chin2020machine}
K.~Chin, T.~Hellebrekers, and C.~Majidi, ``Machine learning for soft robotic
  sensing and control,'' {\em Advanced Intelligent Systems}, vol.~2, no.~6,
  p.~1900171, 2020.

\bibitem{bhirangi2021reskin}
R.~Bhirangi, T.~Hellebrekers, C.~Majidi, and A.~Gupta, ``Reskin: versatile,
  replaceable, lasting tactile skins,'' {\em arXiv preprint arXiv:2111.00071},
  2021.

\bibitem{hellebrekers2019soft}
T.~Hellebrekers, O.~Kroemer, and C.~Majidi, ``Soft magnetic skin for continuous
  deformation sensing,'' {\em Advanced Intelligent Systems}, vol.~1, no.~4,
  p.~1900025, 2019.

\bibitem{mason1985robot}
M.~T. Mason and J.~K. Salisbury~Jr, ``Robot hands and the mechanics of
  manipulation,'' 1985.

\bibitem{bekey1990control}
G.~A. Bekey, R.~Tomovic, and I.~Zeljkovic, ``Control architecture for the
  belgrade/usc hand,'' in {\em Dextrous robot hands}, pp.~136--149, Springer,
  1990.

\bibitem{jacobsen1986design}
S.~Jacobsen, E.~Iversen, D.~Knutti, R.~Johnson, and K.~Biggers, ``Design of the
  utah/mit dextrous hand,'' in {\em Proceedings. 1986 IEEE International
  Conference on Robotics and Automation}, vol.~3, pp.~1520--1532, IEEE, 1986.

\bibitem{iberall1997human}
T.~Iberall, ``Human prehension and dexterous robot hands,'' {\em The
  International Journal of Robotics Research}, vol.~16, no.~3, pp.~285--299,
  1997.

\bibitem{kyberd1994southampton}
P.~J. Kyberd and P.~H. Chappell, ``The southampton hand: an intelligent
  myoelectric prosthesis,'' {\em Journal of rehabilitation Research and
  Development}, vol.~31, no.~4, p.~326, 1994.

\bibitem{tomovic1962adaptive}
R.~Tomovic and G.~Boni, ``An adaptive artificial hand,'' {\em IRE Transactions
  on Automatic Control}, vol.~7, no.~3, pp.~3--10, 1962.

\bibitem{pfeiffer1999shape}
C.~Pfeiffer, K.~DeLaurentis, and C.~Mavroidis, ``Shape memory alloy actuated
  robot prostheses: initial experiments,'' in {\em Proceedings 1999 IEEE
  International Conference on Robotics and Automation (Cat. No. 99CH36288C)},
  vol.~3, pp.~2385--2391, IEEE, 1999.

\bibitem{piazza2019century}
C.~Piazza, G.~Grioli, M.~Catalano, and A.~Bicchi, ``A century of robotic
  hands,'' {\em Annual Review of Control, Robotics, and Autonomous Systems},
  vol.~2, pp.~1--32, 2019.

\bibitem{zhu2019dexterous}
H.~Zhu, A.~Gupta, A.~Rajeswaran, S.~Levine, and V.~Kumar, ``Dexterous
  manipulation with deep reinforcement learning: Efficient, general, and
  low-cost,'' in {\em 2019 International Conference on Robotics and Automation
  (ICRA)}, pp.~3651--3657, IEEE, 2019.

\bibitem{wuthrich2020trifinger}
M.~W{\"u}thrich, F.~Widmaier, F.~Grimminger, J.~Akpo, S.~Joshi, V.~Agrawal,
  B.~Hammoud, M.~Khadiv, M.~Bogdanovic, V.~Berenz, {\em et~al.}, ``Trifinger:
  An open-source robot for learning dexterity,'' {\em arXiv preprint
  arXiv:2008.03596}, 2020.

\bibitem{kumar2016learning}
V.~Kumar, A.~Gupta, E.~Todorov, and S.~Levine, ``Learning dexterous
  manipulation policies from experience and imitation,'' {\em arXiv preprint
  arXiv:1611.05095}, 2016.

\bibitem{nagabandi2020deep}
A.~Nagabandi, K.~Konolige, S.~Levine, and V.~Kumar, ``Deep dynamics models for
  learning dexterous manipulation,'' in {\em Conference on Robot Learning},
  pp.~1101--1112, PMLR, 2020.

\bibitem{kopicki2016one}
M.~Kopicki, R.~Detry, M.~Adjigble, R.~Stolkin, A.~Leonardis, and J.~L. Wyatt,
  ``One-shot learning and generation of dexterous grasps for novel objects,''
  {\em The International Journal of Robotics Research}, vol.~35, no.~8,
  pp.~959--976, 2016.

\bibitem{zeng2020transporter}
A.~Zeng, P.~Florence, J.~Tompson, S.~Welker, J.~Chien, M.~Attarian,
  T.~Armstrong, I.~Krasin, D.~Duong, V.~Sindhwani, {\em et~al.}, ``Transporter
  networks: Rearranging the visual world for robotic manipulation,'' {\em arXiv
  preprint arXiv:2010.14406}, 2020.

\bibitem{shih2020electronic}
B.~Shih, D.~Shah, J.~Li, T.~G. Thuruthel, Y.-L. Park, F.~Iida, Z.~Bao,
  R.~Kramer-Bottiglio, and M.~T. Tolley, ``Electronic skins and machine
  learning for intelligent soft robots,'' {\em Science Robotics}, vol.~5,
  no.~41, p.~eaaz9239, 2020.

\bibitem{cannata2008embedded}
G.~Cannata, M.~Maggiali, G.~Metta, and G.~Sandini, ``An embedded artificial
  skin for humanoid robots,'' in {\em 2008 IEEE International conference on
  multisensor fusion and integration for intelligent systems}, pp.~434--438,
  IEEE, 2008.

\bibitem{hoshi2006large}
T.~Hoshi and H.~Shinoda, ``A large area robot skin based on cell-bridge
  system,'' in {\em SENSORS, 2006 IEEE}, pp.~827--830, IEEE, 2006.

\bibitem{wettels2008biomimetic}
N.~Wettels, V.~J. Santos, R.~S. Johansson, and G.~E. Loeb, ``Biomimetic tactile
  sensor array,'' {\em Advanced Robotics}, vol.~22, pp.~829--849, 2008.

\bibitem{dao2009analysis}
D.~V. Dao, S.~Sugiyama, S.~Hirai, {\em et~al.}, ``Analysis of sliding of a soft
  fingertip embedded with a novel micro force/moment sensor: Simulation,
  experiment, and application,'' in {\em 2009 IEEE International Conference on
  Robotics and Automation}, pp.~889--894, IEEE, 2009.

\bibitem{tomo2017covering}
T.~P. Tomo, A.~Schmitz, W.~K. Wong, H.~Kristanto, S.~Somlor, J.~Hwang,
  L.~Jamone, and S.~Sugano, ``Covering a robot fingertip with uskin: A soft
  electronic skin with distributed 3-axis force sensitive elements for robot
  hands,'' {\em IEEE Robotics and Automation Letters}, vol.~3, no.~1,
  pp.~124--131, 2017.

\bibitem{gandhi2020swoosh}
D.~Gandhi, A.~Gupta, and L.~Pinto, ``Swoosh! rattle! thump!--actions that
  sound,'' {\em arXiv preprint arXiv:2007.01851}, 2020.

\bibitem{clarke2018learning}
S.~Clarke, T.~Rhodes, C.~G. Atkeson, and O.~Kroemer, ``Learning audio feedback
  for estimating amount and flow of granular material,'' {\em Proceedings of
  Machine Learning Research}, vol.~87, 2018.

\bibitem{du2022play}
M.~Du, O.~Y. Lee, S.~Nair, and C.~Finn, ``Play it by ear: Learning skills
  amidst occlusion through audio-visual imitation learning,'' {\em arXiv
  preprint arXiv:2205.14850}, 2022.

\bibitem{hosoda2006anthropomorphic}
K.~Hosoda, Y.~Tada, and M.~Asada, ``Anthropomorphic robotic soft fingertip with
  randomly distributed receptors,'' {\em Robotics and Autonomous Systems},
  vol.~54, no.~2, pp.~104--109, 2006.

\bibitem{van2015learning}
H.~Van~Hoof, T.~Hermans, G.~Neumann, and J.~Peters, ``Learning robot in-hand
  manipulation with tactile features,'' in {\em 2015 IEEE-RAS 15th
  International Conference on Humanoid Robots (Humanoids)}, pp.~121--127, IEEE,
  2015.

\bibitem{odhner2014compliant}
L.~U. Odhner, L.~P. Jentoft, M.~R. Claffee, N.~Corson, Y.~Tenzer, R.~R. Ma,
  M.~Buehler, R.~Kohout, R.~D. Howe, and A.~M. Dollar, ``A compliant,
  underactuated hand for robust manipulation,'' {\em The International Journal
  of Robotics Research}, vol.~33, no.~5, pp.~736--752, 2014.

\bibitem{yuan2017gelsight}
W.~Yuan, S.~Dong, and E.~H. Adelson, ``Gelsight: High-resolution robot tactile
  sensors for estimating geometry and force,'' {\em Sensors}, vol.~17, no.~12,
  p.~2762, 2017.

\bibitem{lambeta2020digit}
M.~Lambeta, P.-W. Chou, S.~Tian, B.~Yang, B.~Maloon, V.~R. Most, D.~Stroud,
  R.~Santos, A.~Byagowi, G.~Kammerer, {\em et~al.}, ``Digit: A novel design for
  a low-cost compact high-resolution tactile sensor with application to in-hand
  manipulation,'' {\em IEEE Robotics and Automation Letters}, vol.~5, no.~3,
  pp.~3838--3845, 2020.

\bibitem{donlon2018gelslim}
E.~Donlon, S.~Dong, M.~Liu, J.~Li, E.~Adelson, and A.~Rodriguez, ``Gelslim: A
  high-resolution, compact, robust, and calibrated tactile-sensing finger,'' in
  {\em 2018 IEEE/RSJ International Conference on Intelligent Robots and Systems
  (IROS)}, pp.~1927--1934, IEEE, 2018.

\bibitem{heyneman2016slip}
B.~Heyneman and M.~R. Cutkosky, ``Slip classification for dynamic tactile array
  sensors,'' {\em The International Journal of Robotics Research}, vol.~35,
  no.~4, pp.~404--421, 2016.

\bibitem{mittendorfer2015realizing}
P.~Mittendorfer, E.~Yoshida, and G.~Cheng, ``Realizing whole-body tactile
  interactions with a self-organizing, multi-modal artificial skin on a
  humanoid robot,'' {\em Advanced Robotics}, vol.~29, no.~1, pp.~51--67, 2015.

\bibitem{sundaram2019learning}
S.~Sundaram, P.~Kellnhofer, Y.~Li, J.-Y. Zhu, A.~Torralba, and W.~Matusik,
  ``Learning the signatures of the human grasp using a scalable tactile
  glove,'' {\em Nature}, vol.~569, no.~7758, pp.~698--702, 2019.

\bibitem{dean2019whole}
E.~Dean-Leon, J.~R. Guadarrama-Olvera, F.~Bergner, and G.~Cheng, ``Whole-body
  active compliance control for humanoid robots with robot skin,'' in {\em 2019
  International Conference on Robotics and Automation (ICRA)}, pp.~5404--5410,
  IEEE, 2019.

\bibitem{funabashi2019morphology}
S.~Funabashi, G.~Yan, A.~Geier, A.~Schmitz, T.~Ogata, and S.~Sugano,
  ``Morphology-specific convolutional neural networks for tactile object
  recognition with a multi-fingered hand,'' in {\em 2019 International
  Conference on Robotics and Automation (ICRA)}, pp.~57--63, IEEE, 2019.

\bibitem{funabashi2022multi}
S.~Funabashi, T.~Isobe, F.~Hongyi, A.~Hiramoto, A.~Schmitz, S.~Sugano, and
  T.~Ogata, ``Multi-fingered in-hand manipulation with various object
  properties using graph convolutional networks and distributed tactile
  sensors,'' {\em IEEE Robotics and Automation Letters}, vol.~7, no.~2,
  pp.~2102--2109, 2022.

\bibitem{tomo2018new}
T.~P. Tomo, M.~Regoli, A.~Schmitz, L.~Natale, H.~Kristanto, S.~Somlor,
  L.~Jamone, G.~Metta, and S.~Sugano, ``A new silicone structure for uskin—a
  soft, distributed, digital 3-axis skin sensor and its integration on the
  humanoid robot icub,'' {\em IEEE Robotics and Automation Letters}, vol.~3,
  no.~3, pp.~2584--2591, 2018.

\bibitem{johansson1996sensory}
R.~S. Johansson, ``Sensory control of dexterous manipulation in humans,'' in
  {\em Hand and brain}, pp.~381--414, Elsevier, 1996.

\bibitem{righetti2014autonomous}
L.~Righetti, M.~Kalakrishnan, P.~Pastor, J.~Binney, J.~Kelly, R.~C. Voorhies,
  G.~S. Sukhatme, and S.~Schaal, ``An autonomous manipulation system based on
  force control and optimization,'' {\em Autonomous Robots}, vol.~36,
  pp.~11--30, 2014.

\bibitem{xia2022review}
Z.~Xia, Z.~Deng, B.~Fang, Y.~Yang, and F.~Sun, ``A review on sensory perception
  for dexterous robotic manipulation,'' {\em International Journal of Advanced
  Robotic Systems}, vol.~19, no.~2, p.~17298806221095974, 2022.

\bibitem{hogan1985impedance}
N.~Hogan, ``Impedance control: An approach to manipulation: Part
  ii—implementation,'' 1985.

\bibitem{mason1981compliance}
M.~T. Mason, ``Compliance and force control for computer controlled
  manipulators,'' {\em IEEE Transactions on Systems, Man, and Cybernetics},
  vol.~11, no.~6, pp.~418--432, 1981.

\bibitem{Polymetis2021}
Y.~Lin, A.~S. Wang, G.~Sutanto, A.~Rai, and F.~Meier, ``Polymetis.''
  \url{https://facebookresearch.github.io/fairo/polymetis/}, 2021.

\bibitem{martinez2017active}
U.~Martinez-Hernandez, T.~J. Dodd, M.~H. Evans, T.~J. Prescott, and N.~F.
  Lepora, ``Active sensorimotor control for tactile exploration,'' {\em
  Robotics and Autonomous Systems}, vol.~87, pp.~15--27, 2017.

\bibitem{yang2015grasp}
Y.~Yang, C.~Fermuller, Y.~Li, and Y.~Aloimonos, ``Grasp type revisited: A
  modern perspective on a classical feature for vision,'' in {\em Proceedings
  of the IEEE conference on computer vision and pattern recognition},
  pp.~400--408, 2015.

\bibitem{feix2015grasp}
T.~Feix, J.~Romero, H.-B. Schmiedmayer, A.~M. Dollar, and D.~Kragic, ``The
  grasp taxonomy of human grasp types,'' {\em IEEE Transactions on
  human-machine systems}, vol.~46, no.~1, pp.~66--77, 2015.

\end{thebibliography}

\newpage
\section*{APPENDIX}




\subsection{Data Collection and Inference}
All training and validation data is collected using the data collection setup described in section \ref{subsec:data-collection}. Data is collected at a frequency of 30 Hz and we collect a baseline measurement before the motor babble policy is executed. To do this, we place the object on the hand and collect 100 sensor measurements, which are then averaged to obtain the baseline measurement. Training and inference works by using baseline-subtracted data

When the models are deployed for bin picking, we use a similar approach to baseline measurement. The robot samples a random location above the bin and moves to this point. We collect 100 sensor measurements and average them to obtain a baseline measurement. The robot then descends and attempts a grasp based on a simple heuristic policy -- the grasp is executed if the norm of the ReSkin data exceeds a threshold. If the grasp succeeds, the trained models are used to infer softness and texture labels from baseline-subtracted data. 

\begin{figure}[h!]
    \centering
    \includegraphics[width=0.8\linewidth]{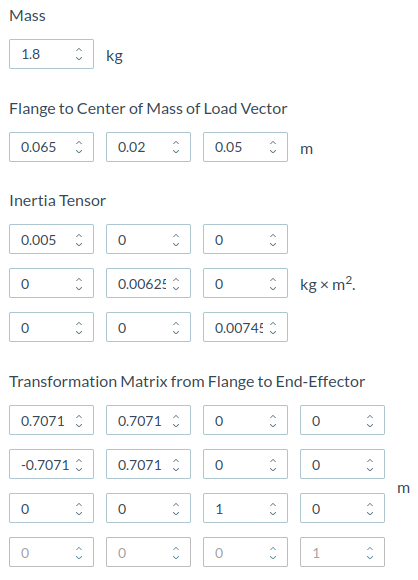}
    \caption{Mass and Inertia matrices for gravity compensation}
    \label{fig:grav-comp}
\end{figure}

The D'Manus weighs 1.5 kilograms (with Reskin sensing). We use Franka Emika robot arm for all our experiments. Figure \ref{fig:grav-comp} outlines the gravity compensation parameters we used. The coordinate frame for these parameters was chosen such that the positive z-axis is normal and pointing outwards from the surface of the palm. The positive x-axis bisects the palm and points from the base of the hand towards the index and little fingers.

\subsection{Ablations}
\label{ablation_details}
We present two sets of ablations: (1) we replace the LSTM in the architecture presented in section \ref{sec:model-learning} with an MLP network that uses stacked frames of ReSkin data to capture temporal correlations, and (2) we train and compare three model variants for the texture classification task described in section \ref{sec:texture-classification}.

\subsubsection{LSTM vs frame-stacking}
\label{app:model-ablation}
\begin{figure}
    \centering
    \includegraphics[width=0.9\linewidth]{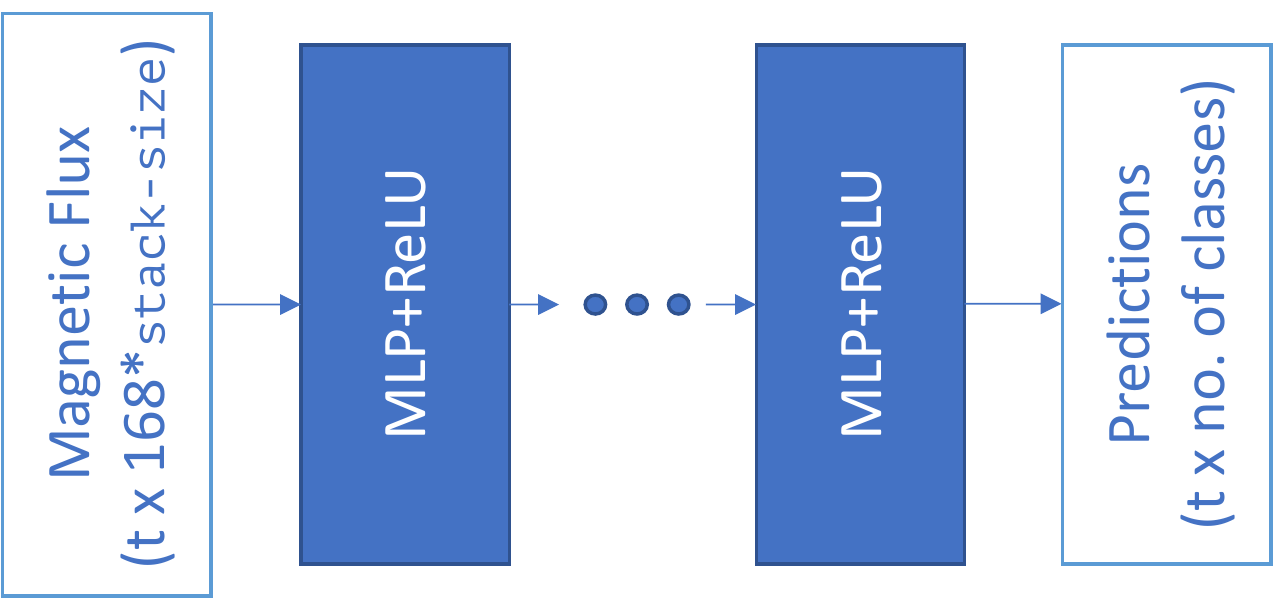}
    \caption{\textit{Frame-stacking} architecture used as an ablation to compare against the LSTM in \ref{fig:model_arch} for capturing temporal correlations.}
    \label{fig:frame-stack}
\end{figure}

We replace the architecture in Fig. \ref{fig:model_arch} with just fully connected network that takes in stacked frames of ReSkin data as input, as shown in Fig. \ref{fig:frame-stack}. We call this the \textit{frame-stacking} architecture. In addition to the number of hidden layers and the size of each layer, this network has two other hyperparameters: \texttt{stack-size}, which refers to the number of frames in the stack of frames that is input to the network, and \texttt{frame-skip}, which refers to the number of frames \textit{skipped} between consecutive frames in the input stack. 

We compare performance of the two architectures for each of the three tasks described in \ref{sec:results}, and present the results in Table \ref{tab:models-comparison}. The results presented in Table \ref{tab:models-comparison} correspond to the best performing models obtained by sweeping over different parameters of the network as detailed in Appendix \ref{app:sweep_params}. We see that for the simpler \textit{Material Identification} task, the \textit{frame-stacking} models performs comparably with the LSTM model. However, as we move to harder tasks like softness and texture classification, we see the LSTM architecture has a significant edge over the \textit{frame-stacking} architecture. 

\begin{table}[h!]
    \centering
    \begin{tabular}{l c}
        \toprule
         Model type & Validation Acc \\
         \midrule 
         \multicolumn{2}{l}{\textit{Material Identification}} \\
         Frame-stacking & 72.00\%\\
         LSTM & 71.24\%\\\midrule
         \multicolumn{2}{l}{\textit{Softness Classification}} \\
         Frame-stacking & 71.38\%\\
         LSTM & 76.17\%\\\midrule
         \multicolumn{2}{l}{\textit{Texture Classification}} \\
         Frame-stacking & 48.85\%\\
         LSTM & 59.03\%\\\bottomrule
    \end{tabular}
    \caption{Comparison of different neural architectures on the object identification, and softness and texture identification tasks}
    \label{tab:models-comparison}
\end{table}

\subsubsection{Texture Classification}
In section \ref{sec:texture-classification}, we elaborate on our choice to train three standalone networks - one for each softness category - for the texture identification task, and using the the prediction of the softness identification network to pick the appropriate network. We term this aggregated model the \textit{Softness-Conditioned} model. In this section, we compare the performance of this model to two other models: (1) vanilla model: a single model with an identical neural architecture, but trained to predict texture category independently of softness, analogous to the softness identification model in \ref{sec:softness-classification}, and (2) joint softness-texture model: a single model that outputs both softness and texture predictions, by minimizing a composite cross-entropy loss, $\mathcal{L} = \mathcal{L}_{ce, softness} + \lambda \mathcal{L}_{ce, texture}$. The goal of training the joint softness-texture model is to allow the network to internally capture dependencies between softness and texture. The results from this comparison are presented in Table \ref{tab:texture-models}. We see that these models do much worse than the \textit{Softness-Conditioned} model, for the reasons described in section \ref{sec:texture-classification}. We train both the \textit{Frame-stacking} and LSTM variants for each of these models. The hyperparameters for these models were also chosen after performing a hyperparameter sweep as described in section \ref{app:sweep_params}.

\begin{table}[h!]
    \centering
    \begin{tabular}{l c}
        \toprule
         Model type & Validation Acc \\
         \midrule 
         \multicolumn{2}{l}{\textit{Softness-Conditioned}} \\
         Frame-stacking & 48.85\%\\
         LSTM & 59.03\%\\\midrule
         \multicolumn{2}{l}{\textit{Vanilla}} \\
         Frame-stacking & 30.43\%\\
         LSTM & 30.67\%\\\midrule
         \multicolumn{2}{l}{\textit{Joint Softness-Texture}} \\
         Frame-stacking & 32.85\%\\
         LSTM & 36.65\%\\\bottomrule
         
    \end{tabular}
    \caption{Comparison of different modeling strategies for texture classification}
    \label{tab:texture-models}
    
\end{table}

\subsection{Parameter Sweeps}
\label{app:sweeps}
We perform two sets of hyperparameter sweeps to improve the performance of our tactile perception models.
\subsubsection{Sweep over Neural Architecture parameters}
\label{app:sweep_params}
All the results presented throughout this paper correspond to the best performing parameters for the LSTM as well as the \textit{Frame-stacking} architectures described in the Appendix. Table \ref{tab:sweep-parameters} outlines the range of hyperparameters we sweep over. Best performing parameters for the LSTM network are described in section \ref{sec:model-learning} and Fig. \ref{fig:model_arch}. For the \textit{Frame-stacking} architecture, we found that 3 hidden layers with 256, 128 and 64 layers and ReLU activations resulted in the best performance. A \texttt{stack-size} of 2, and a \texttt{frame-skip} of up to 5 frames were also sent to be the best stack parameters, as described in \ref{app:model-ablation}.

\begin{table}[h!]
    \centering
    \begin{tabular}{lc}
         Hyperparameter & Sweep Range \\
         \midrule
          \multicolumn{2}{l}{\textbf{Frame-stacking}} \\
         stack-size & [2, 5, 8, 11]\\
         frame-skip & [1, 2, 5, 8, 11]\\
         number of layers & [2, 3, 4] \\
         layer sizes & [64, 128, 256, 512, 1024]\\
         \multicolumn{2}{l}{\textbf{LSTM}} \\
         LSTM: number of layers & [2, 3, 4] \\
         LSTM: layer size & [128, 256, 512, 1024] \\
         fully-connected: number of layers & [1, 2, 3] \\
         fully-connected: layer size & [64, 128, 256, 512] \\\bottomrule
    \end{tabular}
    \caption{Parameter ranges for hyperparameter sweeps}
    \label{tab:sweep-parameters}
\end{table}

\subsubsection{Sweep over Frequency} 
\label{app:sweep_freq}
We test multiple sampling frequencies for the material identification task, and create three datasets with sampling frequencies of 10, 20 and 30 Hz. We then train models, sweeping over the same parameters as described in the last section. Results from this sweep are presented in Table \ref{tab:freq-sweep}. We see that the performance improves significantly as the sampling frequency is increased. Based on this experiment, we stick to a sampling frequency of 30 Hz for all subsequent experiments.

\begin{table}[h!]
    \centering
    \begin{tabular}{cc}
         Frequency & Validation Accuracy \\
         \midrule
         10 Hz & 68.52\%\\
         20 Hz & 79.62\%\\
         30 Hz & 87.04\%\\\bottomrule
    \end{tabular}
    \caption{Validation accuracy over different data collection frequencies for the object identification task}
    \label{tab:freq-sweep}
\end{table}

\end{document}